\documentclass[conference]{IEEEtran}
\IEEEoverridecommandlockouts
\usepackage{cite,multicol}
\usepackage{amsmath,amssymb,amsfonts}

\usepackage{graphicx}
\usepackage{textcomp}
\usepackage{xcolor}
\def\BibTeX{{\rm B\kern-.05em{\sc i\kern-.025em b}\kern-.08em
    T\kern-.1667em\lower.7ex\hbox{E}\kern-.125emX}}

\usepackage{amsmath,amssymb,amsxtra,mathtools}
\usepackage{array}
\usepackage{graphicx}
\usepackage{algorithm}
\usepackage{algpseudocode}
\usepackage{physics}

\usepackage{dsfont,subfigure} 

\usepackage{caption}
\usepackage{verbatim}
\usepackage{enumerate}
\usepackage{parskip}

\usepackage{color}
\definecolor{clemson-orange}{RGB}{234,106,32}
\definecolor{chicago-maroon}{RGB}{128,0,0}
\definecolor{cincinnati-red}{RGB}{190,0,0}
\definecolor{soft-cyan}{RGB}{68,85,90}
\definecolor{firebrick}{RGB}{178,34,34}
\definecolor{crimson}{RGB}{220,20,60}
\definecolor{cerrulean}{rgb}{0.165,0.322,0.745}
\definecolor{jaam}{rgb}{0.45,0.0,0.45}

\usepackage{hyperref}
\hypersetup{
  colorlinks   = true,    % Colours links instead of ugly boxes
  urlcolor     = jaam,    % Colour for external hyperlinks
  linkcolor    = firebrick,    % Colour of internal links
  citecolor    = blue      % Colour of citations
}

\usepackage{parskip}

\newtheorem{theorem}{Theorem}[section]
\newtheorem{lemma}[theorem]{Lemma}

\newtheorem{definition}{Definition}

\newif\ifsolutions \solutionstrue

\def\final{0}
\newcommand{\reviewer}[3]{
  \expandafter\newcommand\csname #1\endcsname[1]{
    \ifthenelse{\equal{\final}{1}} {
      \textcolor{#3}{}
    } {
      \textcolor{#3}{\begin{center} \textbf{#2:} ##1 \end{center}}
    }
  }
}

\reviewer{anirbit}{Anirbit}{jaam}

\renewcommand{\ip}[2]{\left\langle#1,#2\right\rangle}

\newcommand{\ind}[1]{\mathds{1}_{\left\{#1\right\}}}
\newcommand{\relu}{\mathop{\mathrm{ReLU}}}
\newcommand{\sas}{\mathcal{S}\alpha\mathcal{S}}
\newcommand{\distas}{\mathbin{\sim}}

\def\ve#1{\mathchoice{\mbox{\boldmath$\displaystyle\bf#1$}}
{\mbox{\boldmath$\textstyle\bf#1$}}
{\mbox{\boldmath$\scriptstyle\bf#1$}}
{\mbox{\boldmath$\scriptscriptstyle\bf#1$}}}

\newcommand{\x}{{\ve x}}

\newcommand{\z}{{\ve z}}
\newcommand{\g}{{\ve g}}

\newcommand{\w}{{\ve w}}

\newcommand{\A}{\textrm{A}}

\newcommand{\U}{\textrm{U}}

\newcommand{\E}{\mathbb{E}}

\newcommand{\bb}{\mathbb}
\newcommand{\R}{\bb R}

\newcommand{\ignore}[1]{}

\begin{document}

\title{An Empirical Study of the Occurrence of Heavy-Tails in Training a ReLU Gate }

%\thanks{The first author's research is partially supported by NSF-DMS 2124222}

\author{
\IEEEauthorblockN{Sayar Karmakar\IEEEauthorrefmark{1}}
\thanks{\IEEEauthorrefmark{1} 
Department of Statistics, University of Florida, Gainesville, United States,
{\tt sayarkarmakar@ufl.edu}
}
\and 
\IEEEauthorblockN{Anirbit Mukherjee\IEEEauthorrefmark{2}}
\thanks{\IEEEauthorrefmark{2}
Department of Computer Science, University of Manchester, Manchester, United Kingdom,
{\tt anirbit.mukherjee@manchester.ac.uk }
}
}

\maketitle
\begin{abstract}
 A particular direction of recent advance about stochastic deep-learning algorithms has been about uncovering a rather mysterious heavy-tailed nature of the stationary distribution of these algorithms, even when the data distribution is not so. Moreover, the heavy-tail index is known to show interesting dependence on the input dimension of the net, the mini-batch size and the step size of the algorithm. In this short note, we undertake an experimental study of this index for S.G.D. while training a $\relu$ gate (in the realizable and in the binary classification setup) and for a variant of S.G.D. that was proven in \cite{karmakar2022provable} for $\relu$ realizable data. From our experiments we conjecture that these two algorithms have similar heavy-tail behaviour on any data where the later can be proven to converge. Secondly, we demonstrate that the heavy-tail index of the late time iterates in this model scenario has strikingly different properties than either what has been proven for linear hypothesis classes or what has been previously demonstrated for large nets.  
\end{abstract}

\section{Introduction}

The ongoing artificial intelligence revolution \cite{lecun2015deep,sejnowski2020unreasonable,portwood2019turbulence,silver2017mastering,silver2018general} can be said to hinged on developing a vast array of mysterious heuristics which get the neural net to perform ``human like" tasks.  Most such successes can be mathematically seen to be solving the function optimization/``risk minimization" question, $\inf_{\mathbf{N} \in {\cal N}} \mathbb{E}_{\z \in {\cal D}} [\ell (\mathbf{N}, \z)]$ where members of ${\cal N}$ are continuous piecewise linear functions representable by neural nets and $\ell : {\cal N} \times {\rm Support} ({\cal D}) \rightarrow \R^{\geq 0}$ is called a ``loss function" and the algorithm only has sample access to the distribution ${\cal D}$. The successful neural experiments can be seen as suggesting that there are certain special choices of $\ell, {\cal N}$ and ${\cal D}$ for which highly accurate solutions of this seemingly extremely difficult question can be found fast - and often by a surprisingly simple algorithm, Stochastic Gradient Descent. And this is a profound mathematical mystery of our times.  

% Over a parameterized function class (like neural nets) the above question can be reformulated as, $\inf_{\w \in \R^n} \E_{\z \sim {\cal D}} \left [ f(\w, \z) \right ]$ for some appropriate $f$. Then most standard algorithms that are usually used to solve the above, have the following structure, 

% \begin{algorithm2e}
% \caption{A generic stochastic learning algorithm}
% \label{algo:generic}
% \DontPrintSemicolon
% \LinesNumbered
% {\bf Input:} Sampling access to a distribution ${\cal D}$ on $\R^n$\; 
% {\bf Input:} An arbitrarily  or randomly chosen starting point of $\w_1 \in \R^n$ and a constant $\eta > 0$ and a mini-batch size $b$\; 
% \For{$t = 1,\ldots$}{
%     {Sample $\S_t \sim {\cal D}^b$}\;
%     {Form a stochastic (pseudo-)gradient, $\g_t (\w_t,\S_t) \in \R^n$ }\;
%     {\Comment{It would be S.G.D. if $\g_t \coloneqq \frac{1}{b} \sum_{z \in \S_t} \nabla_{\w_t} f(\w_t,\z)$ where the ``$\nabla$" could be a sub-differential.}}\;
%     {$\w_{t+1} := \w_t - \eta \g_t$}\;
% }    
% \end{algorithm2e}

An increasingly popular approach to understand the learning abilities of Stochastic Gradient Descent has been to look at S.D.E.s \cite{raginsky2017non,xu2018global,zhang2017hitting} which can be motivated to be the continuous time limit of S.G.D.s. A key component of this modelling relies on the assumption that the noise in the stochastic gradient is Gaussian in real world learning scenarios. But this assumption has be challenged in recent times in works like \cite{gurbuzbalaban2020heavy} and  \cite{Simsekli:2019:TAIL}. But much remains to be understood about how the tail-index of the noise in the stochastic gradient is at all determined. Also these S.D.E. based results derive the structure of the corresponding stationary distribution by making assumptions about the loss functions which are not verified for neural losses. Hence to make progress it becomes imperative to be able to devise experiments to find out the structure of the stationary distribution in stochastic neural training algorithms.  

On the other hand in works like \cite{jastrzkebski2017three}, albeit motivated by wanting to understand factors influencing the ``width" of the minima found by S.G.D., the authors therein had isolated the important factors influencing the asymptotic behaviour of S.G.D. to be the learning rate (say $\eta$) and the mini-batch size (say $b$). We continue in the same spirit to try to understand the structure of the distribution of the late time iterates of stochastic algorithms which can train a $\relu$ gate. In particular, we focus on the same parameters $\eta$ and $b$ and formulate precise conjectures about how they affect the heavy-tail index of the distribution of asymptotic iterates.   

Via our experiments we will demonstrate that the dependence of this heavy-tail index on $\eta$ and $b$ for training a $\relu$ gate has significant differences from previously reported trends for both linear regression as well as large nets. Thus we motivate targets for future theoretical studies for this model system which can be seen as a sandbox for developing the right tools of study. 

\subsection{Organization} In the subsections \ref{sec:def} and \ref{sec:alg} we will setup the formal definitions that we will use for the heavy-tail index and the specific $\relu$ training algorithms that we will use. In Section \ref{sec:comp} we will give detailed comparison of our approach to previous investigations in the same theme. In Section \ref{sec:sum} we will explain our experimental methodology and summarize our findings. In Section \ref{sec:plot_real} and \ref{sec:plot_bin} we will describe the specific plots that we will obtain and then we conclude with suggestions for future work.    

\subsection{Definitions}\label{sec:def}

Suppose we consider a probability density $p(\x)$ with a power-law tail decreasing as $\frac{1}{\norm{\x}^{\alpha +1}}$ where $\alpha \in (0,2)$. Then for such distributions the $\alpha^{th}-$moment exists for only $\alpha <2$.  Random  variables of such distributions satisfy a central limit theorem too \cite{nolan2003stable} and that has historically motivated the following definition,  

\begin{definition}[Symmetric $\alpha$-Stable Distribution $\sas(\sigma)$]
We write an univariate $X \distas \sas(\sigma)$ if the characteristic function of $X$ is
\[
    \phi_X(t) = \exp(-\sigma^{\alpha}\abs{t}^{\alpha}).
\]
% For a process $X_t$, this means
% \[
%     \phi_{X_t}(u) = \exp\left\{-t\abs{\sigma u}^{\alpha}\right\}.
% \]
\end{definition}

The $\alpha=2$ special case of the above corresponds to Gaussians and the
$\alpha=1$ is the Cauchy distribution. Other values of $\alpha$ in general
do not yield distributions that have nice closed-form
densities. We record the following elementary fact about the $\sas$ distributions. 

\begin{lemma}\label{sas_var}
    If $X\distas \sas(\sigma)$ with $\alpha \ne 2$,
    then the variance of $X$ is infinite.
\end{lemma}

One can further define a class of heavy-tailed distribution that contains the class of Symmetric $\alpha$-Stable distributions.

\begin{definition}[Strictly $\alpha$-Stable Distribution]
$\w$ will be said to be a strictly $\alpha-$stably distributed random vector if for any $m$ independent copies of it say $\w_1,\ldots,\w_m$ we have, 
    $$\sum_{i=1}^m \w_i \overset{\text{in distribution}}{=} m^{\frac {1}{\alpha}} \w.$$ 
\end{definition}

\begin{lemma}\label{lem:ht1ht2}
If $X\distas \sas(\sigma)$, then $X$ is Strictly $\alpha$-Stable. 
\end{lemma}

\begin{IEEEproof}[Proof of Lemma \ref{lem:ht1ht2}]
Assume $X,X_1,\ldots, X_m$ are i.i.d. $\sas(\sigma)$. Then the characteristic function of $\sum_{i=1}^m X_i$ is
\begin{eqnarray*}
\phi_{\sum_{i=1}^m X_i}(t) =& \prod_{i=1}^m \phi_{X_i}(t) = \exp(-\sigma^{\alpha}|m^{1/\alpha}t|^{\alpha})\\
=&\phi_{m^{1/\alpha}X}(t)
\end{eqnarray*}
and thus $\sum_{i=1}^m X_i \overset{\text{in distribution}}{=} m^{\frac {1}{\alpha}} X$. 
\end{IEEEproof}

% \note{ 
% Suppose $X \sim \sas(\sigma)$ then we have, 
% \[ \E \left [ \exp(itX) \right ] = \exp(-\sigma^\alpha \abs{t}^\alpha) \] 
% Hence we have for any constant $k$, 
% \[ \E \left [ \exp(it\cdot kX) \right ] = \exp(-\sigma^\alpha \abs{t\cdot k}^\alpha) \] 
% }

An advantage of using the above larger class of heavy-tailed distributions is that it directly accommodates vector-valued random variables which is the context of this paper. Henceforth, we will only use the later definition and study the estimation of $\alpha$. Interested reader can see \cite{taqqu} for different definitions and variants of the idea of $\alpha$-stable distributions.

\begin{lemma}[{\bf Hill Estimator}, Corollary 2.4 \cite{Mohammadi:2015:EST}]\label{def:hill}
~\\
Let $\{ X_i  \}_{i=1,\ldots,K}$ be a collection of strictly $\alpha-$stably distributed random variables and $K = K_1 \times K_2$. Define $Y_i = \sum_{j=1}^{K_1} X_{j+(i-1)K_1}, \forall i = 1,\ldots,K_2$. Now define the estimator, 
\[ \hat{\frac 1 \alpha} \coloneqq \frac{1}{\log K_1} \cdot \left ( \frac{1}{K_2} \sum_{i=1}^{K_2} \log \vert Y_i \vert - \frac{1}{K} \sum_{i=1}^K \log \vert X_i \vert   \right )    \] 
Then $\hat{\frac 1 \alpha}$ converges almost surely to $\frac{1}{\alpha}$ as $K \rightarrow \infty$. 
\end{lemma}

\subsection{The Algorithms That We Will Study}\label{sec:alg} 

\begin{algorithm}
\caption{Modified mini-batch S.G.D. for training a $\relu$ gate with realizable data}
\label{dadushrelu}
%\LinesNumbered
{\bf Input:} Sampling access to a distribution ${\cal D}_\x$ on $\R^n$ \; \\ 
%, ${\cal D}_\y$ on $\R$ and a threshold value of $\theta_*$.\; \\
{\bf Input:} Oracle access to labels $y \in \R$ when queried with some $\x \in \R^n$\; \\
{\bf Input:} An arbitrarily chosen starting point of $\w_1 \in \R^n$ and a mini-batch size of $b$\;\\ 
{\bf For:} $t = 1,\ldots$\; \\
    \begin{itemize}
        \item Sample i.i.d the data vectors  $ \{\x_{t_1},\ldots,\x_{t_b} \} \sim {\cal D}_{\x}$ and query the oracle with this set.
        \item The Oracle replies with $y_{t_i} = \relu(\ip{\w_*}{\x_{t_i}})$ $\forall i=1,\ldots,b$.
        \item Form the gradient (proxy), 
		\[\g_t := - \frac{1}{b} \sum_{i=1}^b \ind{y_{t_i} > 0} (y_{t_i} - \w_{t}^\top \x_{t_i})\x_{t_i}\]
		\item $\w_{t+1} := \w_t - \eta \g_t$
    \end{itemize} 
\end{algorithm}

For the case of the label of data $\x$ being generated as $y = \relu(\ip{\w_*}{\x})$, in \cite{karmakar2022provable}, it was proven that under mild distributional assumptions, Algorithm \ref{dadushrelu} converges to $\w_*$ in linear time. Such fast convergence for S.G.D. to the global minima on the $\relu$ risk function remains unknown except for Gaussian data.    

We note that the Algorithm \ref{dadushrelu} that we study would have been the standard S.G.D. on the risk,\\ $\E _{\x \sim {\cal D}_{\x}} \left [ \frac{1}{2} \left (  \relu(\ip{\w_*}{\x} - \relu(\ip{\w}{\x}) \right )^2 \right ]$ if in the definition of $\g_t$ in the algorithm the indicators $\ind{y_{t_i} > 0}$ were to be replaced with $\ind{\ip{\w_t}{\x_{t_i}} > 0}$.

%  \ignore{
%  \begin{algorithm}
%  	\caption{\newline Modified mini-batch S.G.D. for training a $\relu$ gate with adversarially perturbed realizable labels.}
%  	\label{dadushrelu:noise}
% 	\begin{algorithmic}[1]
%  		\State {\bf Input:}{Sampling access to a distribution ${\cal D}$ on $\R^n$ and a function $\beta : \R^n \rightarrow [0,1]$}
%  		\State {\bf Input:}{Oracle access to labels $y \in \R$ when queried with some $\x \in \R^n$} 
%  		\State {\bf Input:} An arbitrarily chosen starting point of $\w_1 \in \R^n$ 
%  		\For {$t = 1,\ldots$}{
%  		\State Sample independently $s_t \coloneqq \{\x_{t_1},\ldots,\x_{t_b} \} \sim {\cal D}$ and query the oracle with this set. 
% 		\State The Oracle samples $\forall i=1,\ldots,b, \alpha_{t_i} \sim \{0,1\}$ with probability $\{ 1 -\beta(x_{t_i}), \beta(x_{t_i})\}$
%  		\State The Oracle replies $\forall i=1,\ldots,b, y_{t_i} = \alpha_{t_i} \cdot \xi_{t_i} + \relu(\w_*^{\top}\x_{t_i})$ s.t $\vert \xi_{t_i} \vert \leq \theta_*$ 
		
%  		%where $\xi_{t_i} \distas {\cal D}_\xi$ independently of each other and the $\{\x_{t_1},\ldots,\x_{t_b}\}$
% % 		%\[y_t = f_{\w_*}(\x_t) + \xi_t\]
%  		\State Form the gradient (proxy), 
%  		\[\g_t := - \frac{1}{b} \sum_{i=1}^b \ind{y_{t_i} > \theta_*} (y_{t_i} - \w_t^\top \x_{t_i})\x_{t_i}\]
%  		\State $\w_{t+1} := \w_t - \eta \g_t$
%  		}
%  		%\EndFor 
%  	\end{algorithmic}
%  \end{algorithm}
% }

\section{Comparisons to previous literature on the emergence of heavy-tailed behaviour in neural training}\label{sec:comp}

In a stochastic training algorithm, given a
        mini-batch gradient $\g_t$, and a full gradient $\nabla f$,
        suppose we define the noise vector say $\U = \g_t - \nabla f \in \R^p$. Then
        treating $\U = \{u_1,\ldots, u_p\}$ as a set of $p$ scalars, the authors in \cite{Simsekli:2019:TAIL} had estimated  the tail index of this set i.e., they estimated the tail index of the
        empirical distribution $\frac{1}{p} \sum_{i=1}^p \delta_{u_i}(x)$. We posit that this is not immediately comparable to the project we undertake, that is to measure the tail index of the iterates of the stochastic algorithm.  
        
        The authors in \cite{Simsekli:2019:TAIL} had also concluded from their experiments on fully-connected networks at depths $2$ and $4$ that the mini-batch size does not significantly affect the heavy-tail index that they were measuring. In contrast we will show via our experiments that there are consistent patterns of change w.r.t the mini-batch size of the heavy-tail index of the average late time iterates of S.G.D. 
        
        %Our experiments also demonstrate that the trends here are affected by how the true label generating parameters are sampled between the different S.G.D. samples training the $\relu$ gate on realizable data. 
        
    \cite{gurbuzbalaban2020heavy} trained neural nets using S.G.D. for $10^4$ iterations for $\eta \in [10^{-4},10^{-1}]$
         and $b \in [1,10]$. They treat each layer as a collection of i.i.d. $\alpha-$-stable random variables and measure
the tail-index of each individual layer separately. They observed that, while the
dependence of $\alpha$ on $\frac{\eta}{b}$ differs from layer to layer, in each layer the measured $\alpha$ decreases with increase in the ratio $\frac{\eta}{b}$ in both MNIST and CIFAR. 

But as opposed to the above experiment we will demonstrate that for a single $\relu$ gate the dependence of $\alpha$ on the mini-batch size at a fixed step-length can have much more variations.  Also in our experiments with realizable data we scan the mini-batch dependence of $\alpha$ for $b \in [4,700]$ which is a much larger range of as opposed to the tests done in \cite{gurbuzbalaban2020heavy} for only $b \in [1,10]$

The series of papers\cite{martin2018implicit,martin2019traditional,martin2020heavy} observed an interesting heavy-tailed feature of the power law of the Empirical Spectral Density (ESD) of the weight matrices of deep neural networks. They reported that for larger well trained nets the heavy-tailed property of this ESD becomes more prominent than for smaller networks. It is well-known that the limiting density of eigenvalues of $\A^\top \A$ for a rectangular Gaussian matrix $\A$ follows a Marchenko-Pastur distribution.  They report that such is the distribution of the spectrum of the layer matrices for nets like {\rm LeNet5, MLP3} etc. However, for a diverse array of much larger nets like {\rm AlexNet, Inception} etc. they see that the ESD of the layer matrices is more like that of a matrix $\A^\top\A$ where the entries of $\A$ are samples from a heavy-tailed distribution. They also suggested a multi-phase evolution of the neural training to motivate reasons for this heavy-tail property. This series of papers also reported that for decreasing mini-batch sizes from $500$ to $2$ the E.S.D of fully connected layers of {\rm MiniAlexNet} show more heavy-tailed behavior and while improving generalization accuracy in tandem. 

%Thus they arrive at the conclusion that the large nets, if well-optimized with suitable parameters, will almost universally exhibit heavy-tailed nature.

% \begin{enumerate} 
%         \item Hodgkinson-Mahoney : \cite{hodgkinson2020multiplicative}
%         \item Mirek \cite{Mirek:2010:HEAVY}
%         \item Burac \cite{buraczewski2016stochastic}
% \end{enumerate}

\section{Summary of our experimental results}\label{sec:sum} 
 We shall denote the data as being ``realizable" when the labels are exactly generated from a $\relu$ gate (with unknown weights $\w_*$) i.e $y_{t_i} = \relu(\ip{\w_*}{\x_{t_i}})$ in Algorithm \ref{dadushrelu}. And we recall that whether in Algorithm \ref{dadushrelu} or in S.G.D these algorithms are minimizing the $\ell_2-$risk.  We note that the neural experiments here as well those in \cite{gurbuzbalaban2020heavy}  involve a certain leap of faith. It is because the proof of the Hill estimator being a right measurement of the heavy-tail index uses the assumption that the samples are coming from a strictly $\alpha-$stable distribution - a property which we do not rigorously know to be true for the S.G.D. iterates on a $\relu$ gate or any neural net.  
 
 But it can be observed that via Theorem 4.4.15 of \cite{buraczewski2016stochastic}  it follows that the iterates of the modified $\relu$ training algorithm given in Algorithm \ref{dadushrelu} do have a power-law decay property of the tail for its stationary distribution. {\it And we will demonstrate that all experimental measurements of $\alpha$ on this (mathematically better justified) modified-S.G.D. in Algorithm \ref{dadushrelu} reproduces the standard S.G.D. data within $10 \%$!}

In the experiments in Section \ref{sec:plot_real} when the data is realizable, all measurements are made by invoking the Hill estimator on the {\it average} of the last 1000 centered iterates of S.G.D.s running for 8000-8500 iterations. We do the centering about the true weight $\w_{*,s}$ where the $s$ index is to denote that the global minima is randomly chosen in every sample of S.G.D. that we use. We take $S= K_1 \times K_2$ samples of S.G.D. runs (for every choice of dimension and/or mini-batch size) as needed to match the definition of the Hill estimator in Lemma \ref{def:hill}. We choose $K_1 = K_2 = 25$ motivated by considerations of both accuracy of the Hill estimator as well as trying to keep the run-times manageable. By the $8000^{th}-$iterate the parameter recovery errors i.e $\norm{\w_{t,s} -\w_{*,s}}$ are less than $10^{-6}$ (infact a couple of orders of magnitude lower for most experiments), across all our settings.

Thus, combining the above experimental setup with the context of Algorithm \ref{dadushrelu} we realize that, the distribution whose heavy-tail index we are approximately measuring is the distribution of the random vector, ``$\w_{\infty}(\w_{*,s}) - \w_{*,s}$". (We use the same starting point, the value of $\w_1$, across all samples of the algorithms.) So the measured value of $\alpha$ is affected by the algorithm's parameters like $\eta$ and $b$ as well as how the $\w_{*,s}$ was sampled. We note that this is a feature of the experimental setup of \cite{gurbuzbalaban2020heavy} too. 

As opposed to the experiments in Section \ref{sec:plot_real}, in Section \ref{sec:plot_bin}, we repeat some of the experiments when the data is not realizable. 

From all the experiments, there are $3$ main observations that we present about how the heavy-tails emerge while stochastically training over a $\relu$ gate in both the scenarios of non-realizable data and the case of realizable data (with the $\w_{*,s}$ sampled from a normal distribution).

{\it Firstly,} we see that the heavy-tail index is low (hence the iterates have a more heavy-tailed distribution) for low dimensions and it grows with dimensions. In the realizable setting this growth of $\alpha$ can be conjectured to be sub-linear in dimension.

{\it Secondly}, we note that at any dimension, which is not too low, the tail-index drops with the mini-batch size. 
So larger mini-batches result in heavier tails for the late time iterates of training the $\relu$ gate.   

{\it Thirdly,} we note that wherever tested the properties derived from Algorithm \ref{dadushrelu} and S.G.D. are essentially the same. Given the simplicity of Algorithm \ref{dadushrelu} relative to S.G.D., this observation could potentially be a fulcrum for future theoretical progress in this theme.

% \note{ But looking at the trends it seems to me that one might conjecture that for large enough mini-batches and large enough dimensions its probably approaching alpha = 1 i.e Cauchy! 
% Maybe the right way to build a theory here is to look at a double-scaling limit between dimension and min-batch size and try to show that in this limit $\alpha \to 1$ }

%{\it Note that both the above effects are distinctly different from what was reported in \cite{gurbuzbalaban2020heavy}}

\section{The plots from the regression experiments on realizable data}\label{sec:plot_real}

We present the following specific plots in this section, 

\begin{itemize} 
\item In Figure \ref{fig:G_dim_S.G.D.} we study the input dimension dependence of the heavy-tail index measured on the late time iterates of S.G.D. training a $\relu$ gate (in the realizable) setting when the $\w_{*,s}$s are sampled from a normal distribution. We observe that {\em at higher dimensions the iterates get less heavy-tailed and the growth of $\alpha$ with dimension can be said to be sub-linear.} 

\item In Figure \ref{fig:G_dim_mS.G.D.} we repeat the experiment above for Algorithm \ref{dadushrelu} and check that the trends continue to hold for this too. 

% \item In Figures  \ref{fig:C_dim_S.G.D.} and   \ref{fig:C_dim_mS.G.D.} (in the supplementary) we repeat the above two experiments with $\w_{*,s}$s being sampled from a Cauchy distribution and we see that the pattern holds for a certain range of dimensions.
 
\item In Figure \ref{fig:G_100_S.G.D.} we study the heavy-tail index measured on the late time iterates of S.G.D. training a $\relu$ gate (in the realizable) setting when the $\w_{*,s}$s are sampled from a normal distribution.  We see that beyond a threshold value of the mini-batch size, {\em the distribution of the iterates keep getting heavier tailed as the mini-batch size increases.}  We show that this trend is the same at two different step-lengths separated by an order of magnitude - but the curvature of the $\alpha$ vs $b$ plot in the large mini-batch regime goes from concave to convex as $\eta$ increases. 

\item In Figure \ref{fig:G_100_mS.G.D.} we verify that the general trend as seen for S.G.D. above also continues to hold for the simpler training Algorithm \ref{dadushrelu}. 

%\item In Figure \ref{fig:C_100_S.G.D.} and \ref{fig:C_100_mS.G.D.} (in the supplementary) we repeat the experiments of Figures \ref{fig:G_100_S.G.D.} and \ref{fig:G_100_mS.G.D.} respectively but with the $\w_{*,s}$s being sampled from a Cauchy distribution. This time we see a very different behaviour of how $\alpha$ changes with the mini-batch size whereby beyond a threshold value $\alpha$ seems to be a non-decreasing function of a mini-batch size.  
\end{itemize}

The above $4$ experiments with realizable data, were all done for a $\relu$ gate mapping $\R^{100} \to \R$ and we have checked that the above trends hold for other input dimensions too where we could check within our computational resources. Also all our figures are accompanied by a plot of how the parameter recovery error evolves with time in the corresponding experiment and this acts as a check that our heavy-tail index measurement happens on the iterates when the convergence is fairly complete.

\section{The plots from the experiments in a binary classification setup}\label{sec:plot_bin} 

In here we implement the estimation of $\alpha$ on the average late time iterates of S.G.D. training a $\relu$ gate with $\ell_2-$loss when the labelled data $(\x,y)$ is sampled as follows : firstly, a fair coin is tossed to sample $\x$ from either of two pre-chosen isotropic Gaussian distributions of different variances and whose means are located symmetrically about the origin. Then the corresponding label is assigned to be $1$ or $0$ dependent on which of the two Gaussians the data got sampled from. Although we train on the $\ell_2-$loss, for this case we also report the time evolution of the population risk in the classification loss of the predictor $\x \mapsto \ind{\relu(\ip{\w_t,\x})\geq 0}$ at the $t^{th}-$iterate. To keep the run-times reasonable, in these experiments we are forced to work with much smaller values of input dimension, mini-batch sizes and $K_1 ~\& ~K_2$ than in the previous section. 

\begin{figure*}[!htbp]
    \centering
    %\subfigure[$\eta = 0.005$]
	    {\includegraphics[width=0.9\textwidth,height=4cm]{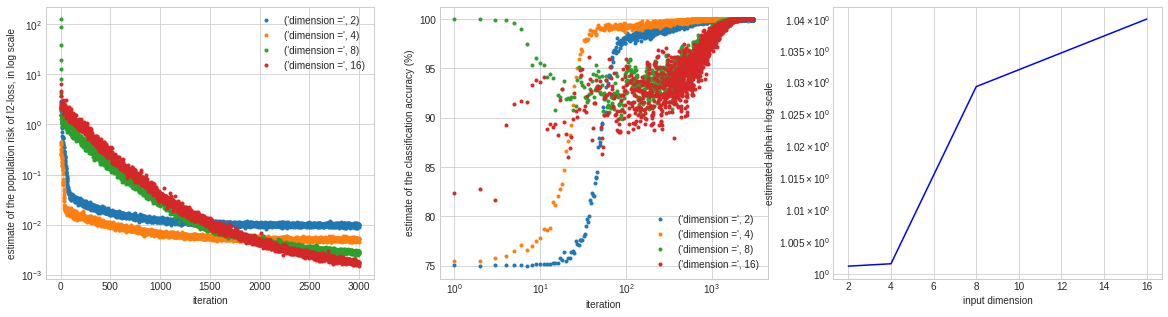}}
	    %\label{fig:C_100_mS.G.D._005}}
% 	\vfill
% 	\subfigure[$\eta = 0.02$]     {\includegraphics[width=0.9\textwidth,height=4cm]{Cauchy_Realizable_modS.G.D._NoNoise_d_100_eta_0_02_alpha_vs_b_GCLT.png}
% 	  \label{fig:C_100_mSGD_02}}
	    \caption{\it Change in the heavy-tail index with input dimension for training a $\relu$ gate via S.G.D. with non-realizable data. }
        \label{fig:Class_alpha_d_S.G.D.} 
\end{figure*}

\begin{figure*}[!htbp]
    \centering
    %\subfigure[$\eta = 0.005$]
	    {\includegraphics[width=0.9\textwidth,height=4cm]{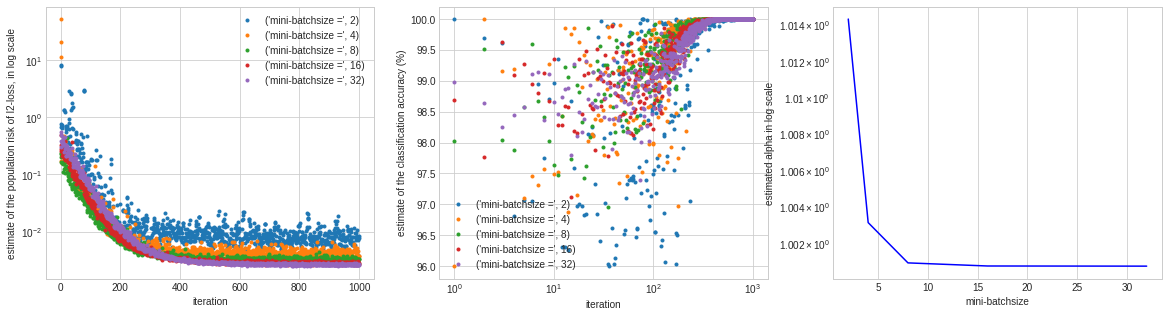}}
	    %\label{fig:C_100_mS.G.D._005}}
% 	\vfill
% 	\subfigure[$\eta = 0.02$]     {\includegraphics[width=0.9\textwidth,height=4cm]{Cauchy_Realizable_modS.G.D._NoNoise_d_100_eta_0_02_alpha_vs_b_GCLT.png}
% 	  \label{fig:C_100_mS.G.D._02}}
	    \caption{\it Change in the heavy-tail index with mini-batch size for training a $\relu$ gate mapping $\R^{8} \to \R$ via S.G.D. with non-realizable data.}
        \label{fig:Class_alpha_b_S.G.D.} 
\end{figure*}

For the above setup we present the following plots, 

\begin{itemize}
    \item In Figure \ref{fig:Class_alpha_d_S.G.D.} we demonstrate how the {\em estimate of $\alpha$ increases with the input dimension of the $\relu$ gate} used. In these experiments the different S.G.D. samples are all initialized from the same point which is chosen from a normal distribution for respective dimensions, and we choose a mini-batch size of $10$ and a step-length of $0.001$. In this case the estimator of Lemma \ref{def:hill} is invoked on samples of the average of the last few hundred iterates when both the regression error as well as the classification error have nearly saturated.   
    
    \item In Figure \ref{fig:Class_alpha_b_S.G.D.} we repeat the same experiment as above but at a fixed input dimension of $8$ and a step-length of $0.005$ and demonstrate how the {\em estimate of $\alpha$ decreases with the mini-batch size used.}
\end{itemize}

\begin{figure}[!htbp]
    \centering
    \subfigure{$\eta = 0.005$}
	{\includegraphics[width = \columnwidth]{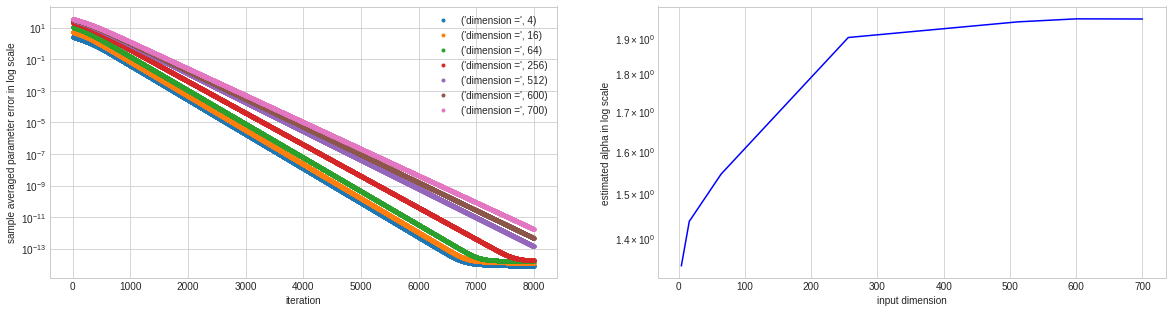}
	 \label{fig:G_dim_S.G.D._005}}
	 %\caption{$\eta = 0.005$}
	%\centering 
	\subfigure{$\eta = 0.01$}    
	{\includegraphics[width = \columnwidth]{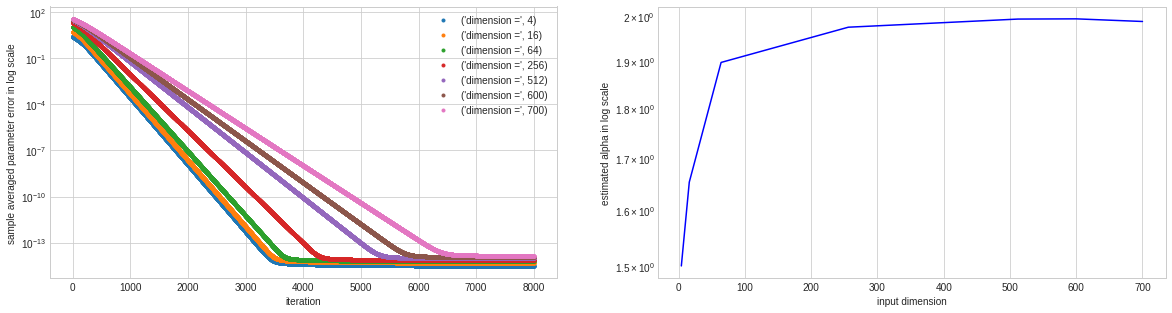}
	 \label{fig:G_dim_S.G.D._01}
	 %\caption{$\eta = 0.01$} 
	}
	%\hspace*{\fill}  
	\caption{Non-decreasing behaviour of the heavy-tail index with input dimension for training a $\relu$ gate in the realizable setting at different step-lengths $\eta$ using S.G.D. }
    \label{fig:G_dim_S.G.D.} 
\end{figure}

\begin{figure}[!htbp]
    \centering
    \subfigure{$\eta = 0.005$}
	{\includegraphics[width = \columnwidth]{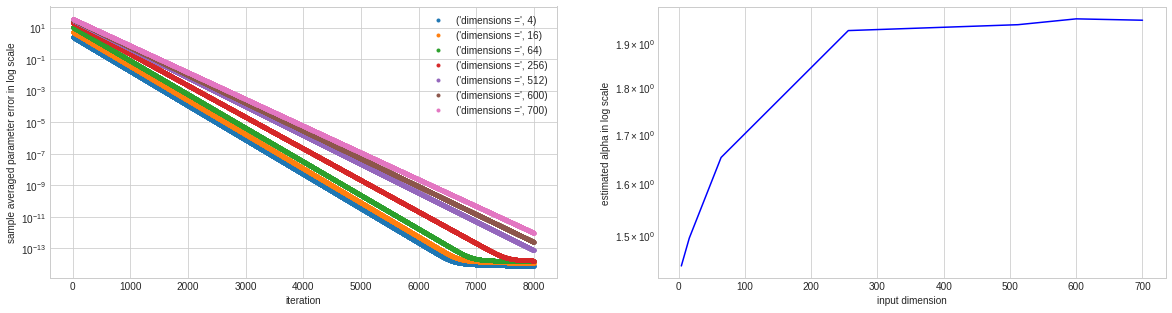}
	 \label{fig:G_dim_S.G.D._005}
	 %\caption{$\eta = 0.005$}
	}
	%\centering 
	\subfigure{$\eta = 0.01$}    
	{\includegraphics[width = \columnwidth]{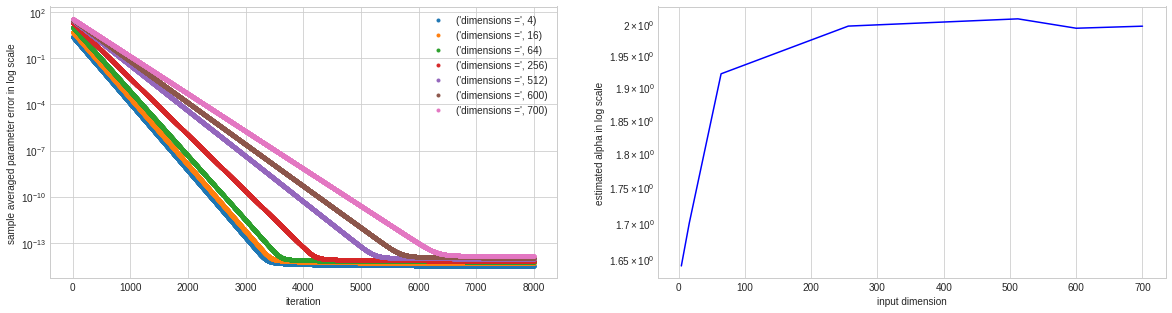}
	 \label{fig:G_dim_S.G.D._01}
	 %\caption{$\eta = 0.01$} 
	}
	%\hspace*{\fill}  
	\caption{Non-decreasing behaviour of the heavy-tail index with input dimension for training a $\relu$ gate in the realizable setting at different step-lengths $\eta$ using {\it Algorithm \ref{dadushrelu}}.}
    \label{fig:G_dim_mS.G.D.} 
\end{figure}

\begin{figure}[!htbp]
    \centering
    \subfigure{$\eta = 0.005$}
	    {\includegraphics[width=\columnwidth]{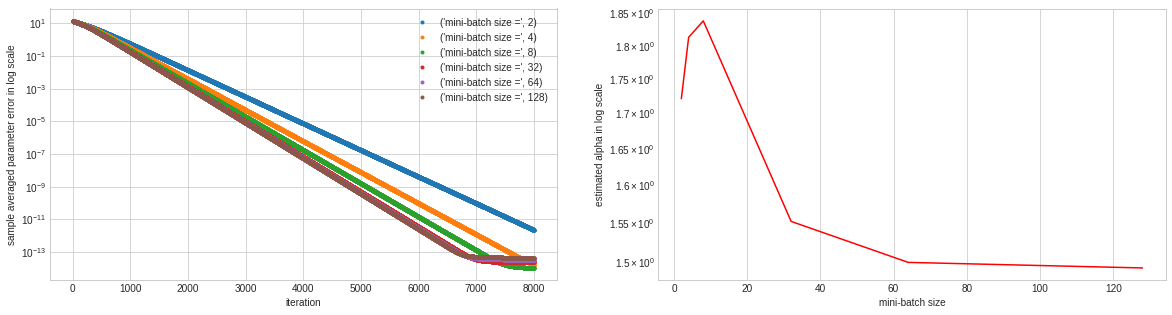}
	    \label{fig:G_100_S.G.D._005}}
	%\vfill
	\subfigure{$\eta = 0.02$}     {\includegraphics[width=\columnwidth]{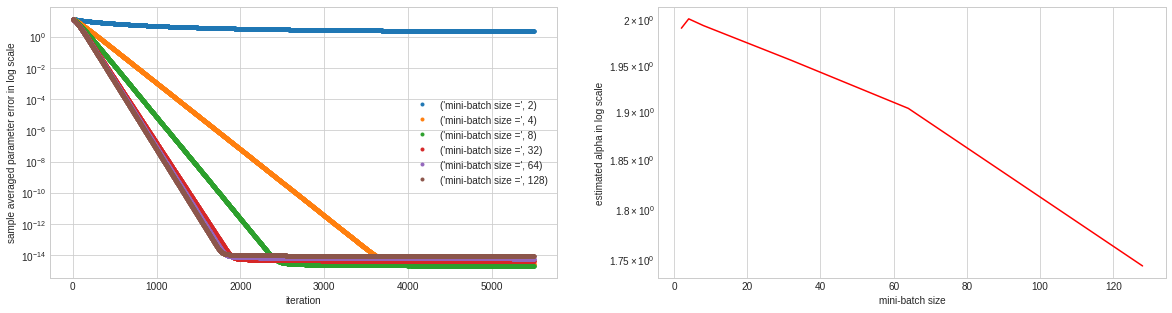}
	  \label{fig:G_100_S.G.D._02}}
	  \caption{{\it Change in the heavy-tail index with mini-batch size} for training a $\relu$ gate in $100$ dimensions in the realizable setting at different step-lengths $\eta$ {\it using S.G.D.} The index measurement is done via using the Hill estimator as described in Section \ref{sec:sum} In each sample of S.G.D. used the generating parameter $\w_*$ was sampled from the standard Gaussian distribution.}
    \label{fig:G_100_S.G.D.} 
\end{figure}

\begin{figure}[!htbp]
    \centering
    \subfigure{$\eta = 0.005$}
	    {\includegraphics[width=\columnwidth]{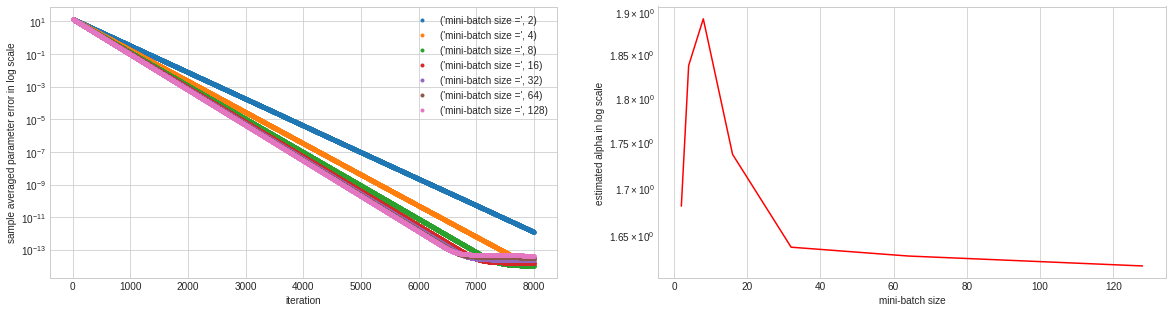}
	    \label{fig:G_100_mS.G.D._005}}
	%\vfill
	\subfigure{$\eta = 0.02$}     {\includegraphics[width=\columnwidth]{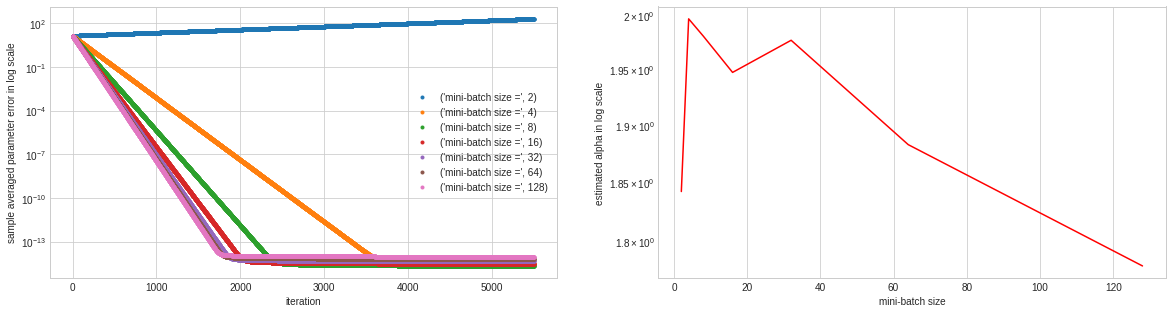}
	    \label{fig:G_100_mS.G.D._02}}
	\caption{{\it Change in the heavy-tail index with mini-batch size} for training a $\relu$ gate in $100$ dimensions in the realizable setting at different step-lengths $\eta$ {\it using   Algorithm \ref{dadushrelu}} . The index measurement is done via using the Hill estimator as described in Section \ref{sec:sum} In each sample of the run of Algorithm  used the generating parameter $\w_*$ was sampled from the standard Gaussian distribution.}
    \label{fig:G_100_mS.G.D.} 
\end{figure}

{\bf Note :} We emphasize that the dependency of $\alpha$ on the input dimension as seen in Figures \ref{fig:G_dim_S.G.D.}, \ref{fig:G_dim_mS.G.D.} and \ref{fig:Class_alpha_d_S.G.D.} and on the mini-batch size $b$ as seen in Figures \ref{fig:G_100_S.G.D.}, \ref{fig:G_100_mS.G.D.} and \ref{fig:Class_alpha_b_S.G.D.}, both differ from the theorems proven for linear regression as well as the experiments done on neural nets in \cite{Gurbuzbalaban:2020:LNREG}.

%\clearpage 
%\section{Experimental results}\label{sec:}

% \begin{figure*}[!htbp]
% \hspace{-8mm}
% \subfloat[Loss function plot for $h=256,n=50$]{
% \includegraphics[scale=0.35]{figures/plot_256_mod.png}
% \label{fig:loss_grad_256}}
% \hspace{-8mm}
% \subfloat[Loss function plot for $h=4096,n=50$]{
% \centering
% \includegraphics[scale=0.35]{figures/plot_4096_mod.png}
% \label{fig:loss_grad_4096}}
% \hspace*{\fill}
% \end{figure*}

%\begin{multicols}{2}
%\end{multicols}

%The index measurement is done via using the Hill estimator as described in Section \ref{sec:sum} In each sample of S.G.D. used the generating parameter $\w_*$ was sampled from the standard Gaussian distribution.

%\clearpage 

%\subsection{Tracking $\alpha$ vs mini-batch size when the global minima is sampled from a Gaussian distribution} 

%\clearpage 
%\paragraph{Tracking $\alpha$ vs mini-batch size when the global minima is sampled from a Cauchy distribution} 
%\vspace{-5cm} 

%\clearpage 
\section{Conclusion and Future work}\label{sec:conclufuture}
Firstly we note that the experiments presented here are very computationally  intensive and repeating these on larger nets or too large input dimensions was beyond our resources, particularly for the case of non-realizable data. Hence an immediate next step would be to devise ways to increase the range of these experiments that were presented here - and to cross-check the index measurements via other estimators \cite{clauset2009power}. 

 Secondly, {\it  in every experiment with realizable data and  $\w_{*,s}$ being sampled from a Gaussian, we have shown that the behaviour of the heavy-tail index for S.G.D. while training on the $\ell_2-$risk on a $\relu$ gate} (recall that $\alpha$ is seen to be non-decreasing w.r.t data dimension \& decreasing w.r.t mini-batch size) {\it  - is very closely reproduced while doing the same training using Algorithm \ref{dadushrelu}} - which is  simpler since it uses updates linear in the weights. We note that the structure of Algorithm \ref{dadushrelu} is more immediately within the ambit of existing theory of stochastic recursions as given in say Theorem 4.4.15 of \cite{buraczewski2016stochastic}.  Hence towards developing a theoretical understanding of the behaviour of the heavy-tail index, we suggest analyzing this property on Algorithm \ref{dadushrelu} as a direction of future research.

\clearpage 

\onecolumn
\bibliographystyle{abbrv}
\bibliography{references}

\begin{appendices}
\section{Code source}
\begin{itemize}
    \item \url{https://colab.research.google.com/drive/1hH-p0CQo3LgvTD4Eq7f0dWeAPenijStE?usp=sharing}
    \item \url{https://colab.research.google.com/drive/1y0S21IXcWpAmX2PzYxMTWVnAePDIQgUH?usp=sharing}
    \item \url{https://colab.research.google.com/drive/1G2PR4k49fTTjkarBgjZ8l93886ovAv1n?usp=sharing}
\end{itemize}
\end{appendices} 
\end{document}

\section{Additional Experiments}
\begin{figure*}[htbp]
    \centering
    \subfigure[$\eta = 0.005$]
	    {\includegraphics[width=0.9\textwidth,height=4cm]{cauchyS.G.D.final2.png}
	    \label{fig:C_dim_S.G.D._005}}
	\vfill
	\subfigure[$\eta = 0.01$]     {\includegraphics[width=0.9\textwidth,height=4cm]{cauchyS.G.D.final1.png}
	  \label{fig:C_dim_S.G.D._01}}
	    \caption{Change in the heavy-tail index with input dimension for training a $\relu$ gate in the realizable setting at different step-lengths $\eta$ using S.G.D. The index measurement is done via using the Hill estimator as described in Section \ref{sec:sum} In each sample of S.G.D. used the generating parameter $\w_*$ was sampled from the standard Cauchy distribution.}
        \label{fig:C_dim_S.G.D.} 
\end{figure*}

\begin{figure*}[htbp]
    \centering
    \subfigure[$\eta = 0.005$]
	    {\includegraphics[width=0.9\textwidth,height=4cm]{cauchymodS.G.D.final2.png}
	    \label{fig:C_dim_mS.G.D._005}}
	\vfill
	\subfigure[$\eta = 0.01$]     {\includegraphics[width=0.9\textwidth,height=4cm]{cauchymodS.G.D.final1.png}
	  \label{fig:C_dim_mS.G.D._01}}
	    \caption{Change in the heavy-tail index with input dimension for training a $\relu$ gate in the realizable setting at different step-lengths $\eta$ using Algorithm \ref{dadushrelu}. The index measurement is done via using the Hill estimator as described in Section \ref{sec:sum} In each sample of S.G.D. used the generating parameter $\w_*$ was sampled from the standard Cauchy distribution.}
        \label{fig:C_dim_mS.G.D.} 
\end{figure*}

\begin{figure*}[t]
    \centering
    \subfigure[$\eta = 0.005$]
	    {\includegraphics[width=0.9\textwidth,height=4cm]{Cauchy_Realizable_S.G.D._NoNoise_d_100_eta_0_005_alpha_vs_b_GCLT.png}
	    \label{fig:C_100_S.G.D._005}}
	\vfill
	\subfigure[$\eta = 0.05$]     {\includegraphics[width=0.9\textwidth,height=4cm]{Cauchy_Realizable_S.G.D._NoNoise_d_100_eta_0_05_alpha_vs_b_GCLT.png}
	  \label{fig:C_100_S.G.D._05}}
	    \caption{Change in the heavy-tail index with mini-batch size for training a $\relu$ gate in $100$ dimensions in the realizable setting at different step-lengths $\eta$ using S.G.D. The index measurement is done via using the Hill estimator as described in Section \ref{sec:sum} In each sample of S.G.D. used the generating parameter $\w_*$ was sampled from the standard Cauchy distribution.}
        \label{fig:C_100_S.G.D.} 
\end{figure*}

\begin{figure*}[t]
    \centering
    \subfigure[$\eta = 0.005$]
	    {\includegraphics[width=0.9\textwidth,height=4cm]{Cauchy_Realizable_modS.G.D._NoNoise_d_100_eta_0_005_alpha_vs_b_GCLT.png}
	    \label{fig:C_100_mS.G.D._005}}
	\vfill
	\subfigure[$\eta = 0.02$]     {\includegraphics[width=0.9\textwidth,height=4cm]{Cauchy_Realizable_modS.G.D._NoNoise_d_100_eta_0_02_alpha_vs_b_GCLT.png}
	  \label{fig:C_100_mS.G.D._02}}
	    \caption{Change in the heavy-tail index with mini-batch size for training a $\relu$ gate in $100$ dimensions in the realizable setting at different step-lengths $\eta$ using ``linearized S.G.D." as given in Algorithm \ref{dadushrelu}. The index measurement is done via using the Hill estimator as described in Section \ref{sec:sum} In each sample of the run of Algorithm  used the generating parameter $\w_*$ was sampled from the standard Cauchy distribution.}
        \label{fig:C_100_mS.G.D.} 
\end{figure*}

%\end{document}
\end{document}